\documentclass[fleqn,10pt]{wlscirep}
\usepackage[utf8]{inputenc}
\usepackage[T1]{fontenc}
\usepackage{algorithmic}
\usepackage[ruled]{algorithm2e}
\usepackage{bm}
\usepackage{chngcntr}

\title{BRIGHT: A Collaborative Generalist-Specialist Foundation Model for Breast Pathology}
\author[1,$\dagger$,*]{Xiaojing Guo} 
\author[2,3,$\dagger$]{Jiatai Lin} 
\author[1,$\dagger$]{Yumian Jia} 
\author[5,$\dagger$]{Jingqi Huang} 
\author[4,$\dagger$]{Zeyan Xu} 
\author[1,$\dagger$]{Weidong Li} 

\author[5]{Longfei Wang} 
\author[6]{Jingjing Chen} 
\author[7]{Qin Li} 
\author[8]{Weiwei Wang} 
\author[9]{Lifang Cui} 
\author[10]{Wen Yue} 
\author[11]{Zhiqiang Cheng} 
\author[12]{Xiaolong Wei} 
\author[13]{Jianzhong Yu} 
\author[14]{Xia Jin} 
\author[15,16]{Baizhou Li} 
\author[17]{Honghong Shen} 
\author[18]{Jing Li} 
\author[19]{Chunlan Li} 
\author[20]{Yanfen Cui} 
\author[21]{Yi Dai} 

\author[1]{Yiling Yang} 
\author[1]{Xiaolong Qian} 
\author[1]{Liu Yang} 
\author[1]{Yang Yang} 
\author[1]{Guangshen Gao} 
\author[1]{Yaqing Li} 
\author[1]{Lili Zhai} 
\author[1]{Chenying Liu} 
\author[1]{Tianhua Zhang} 
\author[2,3]{Zhenwei Shi} 
\author[2,3]{Cheng Lu} 
\author[22]{Xingchen Zhou} 
\author[23]{Jing Xu} 

\author[24,*]{Miaoqing Zhao} 
\author[25,*]{Fang Mei} 
\author[26,27,*]{Jiaojiao Zhou} 
\author[28,29,30,31,*]{Ning Mao} 
\author[1,*]{Fangfang Liu} 
\author[2,3,4,*]{Chu Han} 
\author[2,3,4,*]{Zaiyi Liu}

\affil[1]{Department of Breast Pathology and Laboratory, Tianjin Medical University Cancer Institute \& Hospital, National Clinical Research Center for Cancer, Key Laboratory of Breast Cancer Prevention and Therapy, Tianjin Medical University, Ministry of Education, Tianjin’s Clinical Research Center for Cancer, West Huanhu Road, Tianjin, China}
\affil[2]{Guangdong Provincial Key Laboratory of Artificial Intelligence in Medical Image Analysis and Application, Guangdong Provincial People's Hospital (Guangdong Academy of Medical Sciences), Southern Medical University, Guangzhou, China}
\affil[3]{Department of Radiology, Guangdong Provincial People's Hospital (Guangdong Academy of Medical Sciences), Southern Medical University, Guangzhou, China}
\affil[4]{Department of Radiology, Yunnan Cancer Hospital, The Third Affiliated Hospital of Kunming Medical University, Peking University Cancer Hospital Yunnan, Yunnan, China}
\affil[5]{School of Biomedical Engineering, Southern Medical University, Guangzhou, China}
\affil[6]{Department of Radiology, Qingdao University Hospital, Qingdao, Shandong, China}
\affil[7]{Department of Medical Imaging, Weifang Hospital of Traditional Chinese Medicine, Weifang, Shandong, China}
\affil[8]{Department of Medical Imaging, Affiliated Hospital of Jining Medical University, Jining, Shandong, China}
\affil[9]{{Department of Pathology}, China-Japan Friendship Hospital, {Beijing}, China}
\affil[10]{Department of Pathology, The Obstetrics \& Gynecology Hospital of FuDan University, Shanghai, China}
\affil[11]{Department of Pathology, ShenZhen Third People's Hospital, Guangdong, China}
\affil[12]{Department of Pathology, Cancer Hospital of ShanTou University Medical College, Guangdong, China}
\affil[13]{Department of Pathology, DongGuan People's Hospital, Guangdong, China}
\affil[14]{Department of Pathology, Hunan Cancer Hospital, The Affiliated Cancer Hospital of Xiangya School of Medicine, Central South University, Hunan, China}
\affil[15]{Department of Pathology, The Second Affiliated Hospital Zhejiang University School of Medicine, Zhejiang, China}
\affil[16]{{Department of Pathology}, The Fourth Affiliated Hospital, Zhejiang University School of Medicine, Zhejiang, China}
\affil[17]{Department of Pathology, Second hospital of ShanXi Medical University, ShanXi, China}
\affil[18]{Department of Pathology, ZiBo Central Hospital, Shandong, China}
\affil[19]{Department of Pathology, QiLu Hospital of Shandong University, Shandong, China}

\affil[20]{Department of Radiology, Shanxi Province Cancer Hospital, Shanxi Hospital Affiliated to Cancer Hospital, Chinese Academy of Medical Sciences/Cancer Hospital Affiliated to Shanxi Medical University, Shanxi, China}
\affil[21]{Department of Medical Imaging, Peking University Shenzhen Hospital, Guangdong, China}
\affil[22]{Department of Pathology, The Second Hospital, Cheeloo College of Medicine, Shandong University}
\affil[23]{Department of Pathology, Qingdao Central Hospital, University of Health and Rehabilitation Sciences, Qingdao, China}

\affil[24]{Department of Pathology,  Shandong First Medical University Affiliated Cancer Hospital, Shandong, China}
\affil[25]
{Department of Pathology, Peking University Third Hospital, School of Basic Medical Sciences, Peking University Health Science Center, Beijing, China}
\affil[26]{Department of Breast Surgery and Oncology, the Second Affiliated Hospital, Zhejiang University School of Medicine, Hangzhou, China}
\affil[27]{The Key Laboratory of Cancer Prevention and Intervention, China}

\affil[28]{Big Data and Artificial Intelligence Laboratory, Yantai Yuhuangding Hospital, Qingdao University, Yantai, Shandong, China}
\affil[29]{Department of Radiology, Yantai Yuhuangding Hospital, Qingdao University, Yantai, Shandong, China}
\affil[30]{Shandong Provincial Key Medical and Health Laboratory of Intelligent Diagnosis and Treatment for Women’s Diseases, Yantai Yuhuangding Hospital, Qingdao University, Yantai, Shandong, China}
\affil[31]{Faculty of Applied Sciences, Macao Polytechnic University, Macao, China}

\affil[$\dagger$]{These authors contributed equally to this work.}
\affil[*]{Correspondence: guoxiaojing@tjmuch.com (Xiaojing Guo), zhaomqsd@163.com (Miaoqing Zhao), meifangmail@bjmu.edu.cn (Fang Mei), zhoujj@zju.edu.cn (Jiaojiao Zhou), {maoning@pku.edu.cn} (Ning Mao), {liufangfang@tjmuch.com} (Fangfang Liu), {hanchu@gdph.org.cn} (Chu Han), {liuzaiyi@gdph.org.cn} (Zaiyi Liu)}

\begin{abstract}
Generalist pathology foundation models (PFMs), pretrained on large-scale multi-organ datasets, have demonstrated remarkable predictive capabilities across diverse clinical applications. However, their proficiency on the full spectrum of clinically essential tasks within a specific organ system remains an open question due to the lack of large-scale validation cohorts for a single organ as well as the absence of a tailored training paradigm that can effectively translate broad histomorphological knowledge into the organ-specific expertise required for specialist-level interpretation. In this study, we propose BRIGHT, the first PFM specifically designed for breast pathology, trained on over 51,000 breast whole-slide images derived from a cohort of over 40,000 patients across 19 hospitals. BRIGHT employs a collaborative generalist-specialist framework to capture both universal and organ-specific features. To comprehensively evaluate the performance of PFMs on breast oncology, we curate the largest multi-institutional cohorts to date for downstream task development and evaluation, comprising over 25,000 WSIs across 10 hospitals. The validation cohorts cover the full spectrum of breast pathology across 25 distinct clinical tasks spanning diagnosis, biomarker prediction, treatment response and survival prediction. Extensive experiments demonstrate that BRIGHT outperforms five leading generalist PFMs, achieving state-of-the-art (SOTA) performance in 25 of 25 internal validation tasks and in 4 of 11 external validation tasks with excellent heatmap interpretability. By evaluating on large-scale validation cohorts, this study not only demonstrates BRIGHT's clinical utility in breast oncology but also validates a collaborative generalist-specialist paradigm, providing a scalable template for developing PFMs on a specific organ system, accelerating the translation of foundation models into specialized clinical practice.
\end{abstract}
\begin{document}

\flushbottom
\maketitle

\thispagestyle{empty}
\section*{KEYWORDS}
Computational pathology, Foundation model, Breast cancer, Generalist-specialist collaboration


\section*{Introduction}
\label{sec:Introduction}
The advent of computational pathology (CPath)~\cite{marra2025artificial}, powered by deep learning~\cite{lecun2015deep}, holds transformative potential for oncology. By extracting sub-visual features from whole-slide images (WSIs), CPath techniques promise to augment pathological diagnosis~\cite{lu2021ai,bulten2022artificial}, predict molecular profiles~\cite{wagner2023transformer,campanella2025real,gustav2025assessing}, treatment response~\cite{ogier2023federated} and patient prognosis~\cite{yuan2025pancancer,skrede2020deep,amgad2024population}, thereby advancing the goals of precision medicine. However, the clinical translation of these models has been largely constrained by a dominant paradigm of developing narrow, task-specific models. They are typically trained on limited, single-institution datasets for a singular clinical endpoint, resulting in unsatisfactory performance and poor generalization across diverse clinical settings.

Given the pressing need of scalable AI in pathology, a new paradigm of Pathology Foundation Models (PFMs) has emerged. Inspired by breakthroughs in computer vision, recent studies have adapted state-of-the-art self-supervised learning (SSL) algorithms~\cite{oquab2023dinov2,radford2021learning} to train models on vast and diverse multi-organ WSI datasets~\cite{moor2023foundation,huang2023visual,wang2024pathology,lu2024visual,xiang2025vision,lu2024multimodal,xu2024whole,chen2024towards,vorontsov2024foundation,yang2025foundation,ma2025generalizable}. This pre-training strategy allows a model to learn a universal and transferable representation of fundamental histopathological patterns without any manual annotation~\cite{xiong2025survey}. Once established, such a generalist PFM can be efficiently adapted with minimal labeled data to a wide array of downstream clinical tasks. Furthermore, the benefits also include robust performance in data-limited (few-shot) settings, zero-shot capabilities, and significantly faster convergence during task-specific training compared to models trained from scratch~\cite{chen2024towards}.

Despite their demonstrated versatility, the capability and generalizability of existing generalist PFMs when applied to the comprehensive clinical workflow of a single, specialized organ system remain unclear for the following two reasons. From the benchmarking perspective, the field of CPath is actually constrained by a severe scarcity of large-scale, open-source pathology datasets~\cite{mahmood2025benchmarking}. Compounded by the inherent difficulty of assembling sufficient private data for a specific organ, most generalist PFM studies evaluated their organ-specific performance on limited public datasets. Consequently, these evaluations rely on data that are typically small in scale, lack the heterogeneity of real-world clinical practice and finally fail to encompass the full spectrum of diagnostic, prognostic and predictive tasks required for comprehensive organ-specific assessment. Even dedicated PFM benchmarking studies currently lack the necessary organ-focused, task-comprehensive cohorts~\cite{ma2025pathbench,neidlinger2025benchmarking,campanella2025clinical}. Such a design measures multi-organ versatility rather than depth of expertise within a single domain, leaving a critical gap from generalist to specialist. From the technical perspective, current generalist PFMs are primarily developed using multi-organ datasets. While this strategy effectively learns universal histomorphological patterns, it might fail to represent the organ-specific expertise features that are critical for specialist-level interpretation within a single organ system. This limitation is further corroborated by evidence suggesting that different PFMs, pretrained with distinct strategies on varied datasets, exhibit specialized advantages on different evaluation tasks~\cite{ma2025generalizable}.

\begin{figure}
	\centering
	\includegraphics[width=.985\linewidth]{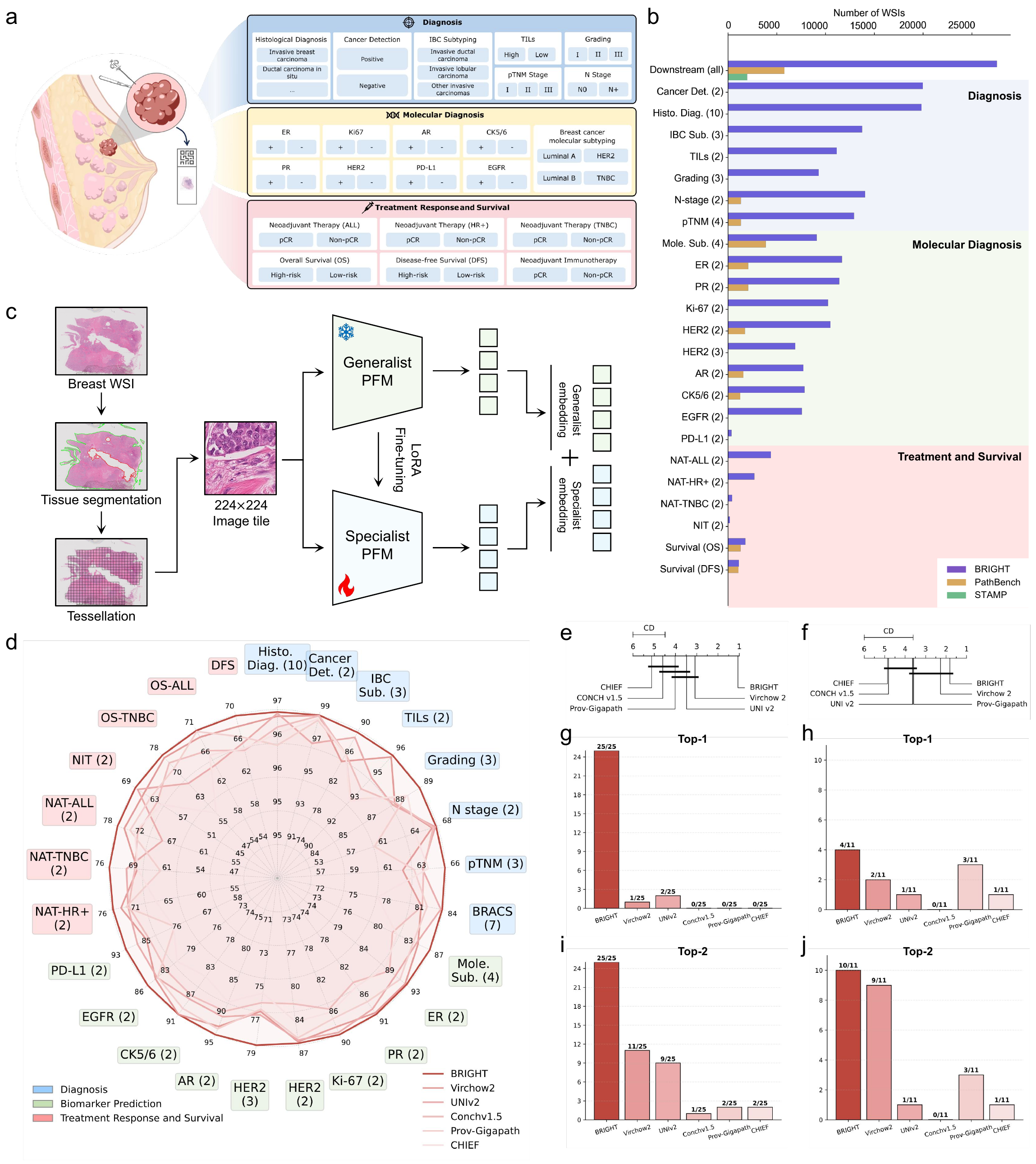}
	\caption{\textbf{Overview of BRIGHT study.} \textbf{a}, Schematic of the clinical spectrum covered by the study, encompassing diagnostic subtyping, molecular biomarker prediction, treatment response assessment, and long-term survival prediction in breast oncology. \textbf{b}, Distribution of datasets used for the development and evaluation of each downstream clinical prediction task. BRIGHT covers wider clinical tasks and more benchmarking data than existing benchmarking studies. \textbf{c}, Architecture of BRIGHT. A leading generalist pathology foundation model (PFM, Virchow2) is specialized for breast pathology via Low-Rank Adaptation (LoRA), yielding a specialist PFM (BRIGHT (S)). The final BRIGHT model is formed by the collaborative integration of feature embeddings from both generalist and specialist encoders. \textbf{d}, BRIGHT generally outperforms five leading PFMs in 25 internal validation benchmarks. {\textbf{e-f}, Critical diagram~(CD) for internal and external test. \textbf{g-j}, Model ranking summary. The number of tasks (among 25 internal \textbf{(g,i)} and 11 external \textbf{(h,j)} benchmarks) in which each foundation model achieved top-1 (\textbf{g} and \textbf{h} for internal and external test) or top-2 (\textbf{i} and \textbf{j} for internal and external test) performance based on AUROC for classification tasks and C-index for survival prediction tasks.}}
	\label{fig1}
\end{figure}

In this study, we introduce BRIGHT (\textbf{BR}east pat\textbf{H}olo\textbf{G}y founda\textbf{TI}on model), a collaborative generalist-specialist framework designed to establish the first PFM for breast pathology. Since existing PFMs have already learned the generalist histomorphological patterns very well. The core idea of BRIGHT is to integrate the well-learned generalist features with the specialist organ-specific features learned from breast pathology data. We first train a specialist model BRIGHT (S) from a leading PFM Virchow2~\cite{vorontsov2024foundation,zimmermann2024virchow2} that trained on over 3 million WSIs, using an efficient and parameter-effective fine-tuning strategy low-rank adaptation (LoRA)~\cite{hu2022lora} to adapt the generalist foundation to the specialized domain of breast pathology (Figure~\ref{fig1}c). Then we integrate the feature representations by collaboratively fusing the generalist feature embeddings extracted from the original Virchow2 encoder with the specialist feature embeddings extracted from the adapted BRIGHT (S) encoder. This collaborative generalist-specialist architecture enables BRIGHT to simultaneously leverage broad histomorphological understanding and nuanced, breast-specific diagnostic insights.

BRIGHT is pre-trained on approximately 210 million histopathology tiles extracted from 51,836 breast WSIs (Figure~\ref{fig1}c), including 42,876 WSIs sourced from a five-year, consecutively collected data of 32,054 patients from our primary cohort (C1.TMUCIH) and 8,960 WSIs across 18 major institutions. As a result, both training and validation cohorts comprehensively cover various breast disease types, spanning almost the entire spectrum from benign lesions, in situ carcinomas, to invasive carcinomas of various histological and molecular subtypes. It ensures that BRIGHT learns a rich and diverse feature representation of breast-specific pathology and provides a robust basis for evaluating the real-world clinical performance of PFMs. To thoroughly evaluate BRIGHT and the broader capabilities of PFMs in breast oncology, we have established a comprehensive benchmarking dataset comprising 25 clinically essential tasks. These tasks span the full clinical workflow, including diagnosis, biomarker prediction, treatment response assessment, and survival prediction (Figure~\ref{fig1}a). The development and validation of the downstream models are supported by the cohort with over 25,000 of WSIs which is much larger than existing PFM benchmarking studies (Figure~\ref{fig1}b). This substantial scale ensures more robust and stable model training than that trained on limited samples, providing a reliable reflection of real-world clinical performance. Furthermore, the size and diversity of these cohorts enable meaningful subgroup modeling to evaluate predictive performance across different treatment strategies for different molecular subtypes.

We comprehensively evaluate BRIGHT across 25 benchmarks (Figure~\ref{fig1}d) covering the full spectrum of the clinical applications. BRIGHT generally outperforms five existing generalist PFMs, reflecting the efficacy of the collaborative generalist-specialist design. {Notably, BRIGHT achieves top-1/top-2 (or tied for top-1/top-2) performance in 25 (25) out of 25 internal tasks (Figure~\ref{fig1}g, i)) and 4 (10) out of 11 external tasks (Figure~\ref{fig1}h, j) based on statistical significance. Two critical difference (CD) diagrams based on Nemenyi test demonstrate the advantage of our method over several PFM baselines and the corresponding critical difference intervals for internal~(Figure~\ref{fig1}e) and external~(Figure~\ref{fig1}f) test.}
These findings highlight BRIGHT not only as a state-of-the-art model for breast oncology but also as a scalable template for developing high-fidelity, organ-specific foundation models, thereby accelerating the translation of artificial intelligence into specialized clinical practice. 


\section*{Results}
The BRIGHT model was pre-trained on a large-scale, diverse cohort of breast WSIs, encompassing the full pathological spectrum from benign lesions and in situ carcinomas to invasive carcinomas of varied histological and molecular subtypes. This comprehensive training enables BRIGHT to learn a rich and diverse feature representation of breast-specific pathology. To rigorously assess its clinical utility, we evaluated BRIGHT-derived feature embeddings across 25 distinct downstream tasks, spanning four major clinical domains, including histological diagnosis, biomarker prediction, treatment response prediction and prognosis prediction. Following the established weakly-supervised paradigm in computational pathology, we employed a multiple-instance learning (MIL) framework CLAM~\cite{lu2021data} for task-specific downstream adaptation. For a robust internal evaluation, we constructed a held-out validation cohort comprising all consecutive patients undergoing breast biopsy from the full calendar year of 2024 at our primary institution (C1.TMUCIH), ensuring a natural and unselected distribution reflective of real-world breast oncology practice. To evaluate the generalizability of BRIGHT, external validation is conducted on the WSIs collected from 9 external hospitals. To ensure a rigorous and unbiased evaluation, all WSIs used for downstream task assessment were held-out from the BRIGHT foundation model's pre-training phase. The detailed dataset distribution of each task is shown in Supplementary Table S1 (primary cohort) and S2 (external cohorts). We compare BRIGHT with five leading PFMs, including four image-based PFMs (Virchow2~\cite{vorontsov2024foundation,zimmermann2024virchow2}, UNIv2~\cite{chen2024towards}, {Prov-GigaPath~\cite{xu2024whole} and CHIEF~\cite{wang2024pathology})}, one visual-language PFM (CONCHv1.5~\cite{lu2024visual}).

\subsection*{BRIGHT can enhance diagnostic accuracy}
\begin{figure}[htp]
	\centering
	\includegraphics[width=.985\linewidth]{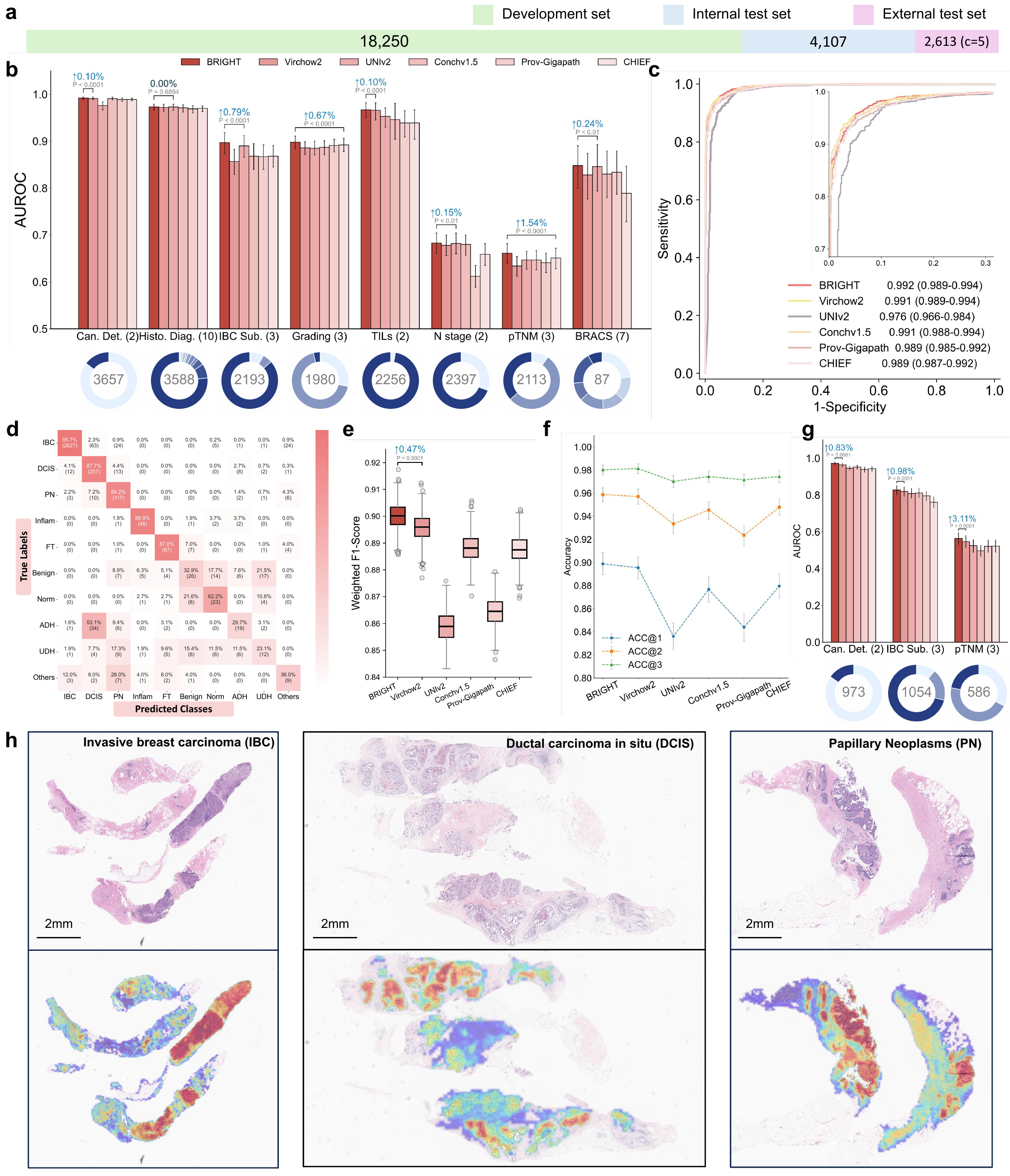}
	\caption{\textbf{Performance of PFMs on diagnostic tasks of breast pathology.} \textbf{a}, Distribution of the downstream model development, internal test and external test datasets. The number in brackets is the number of hospitals for external validation. \textbf{b}, Model performance (AUROC) across individual diagnostic tasks on the internal validation set. The pie charts below indicate the total number of WSIs and the class distribution for each task. The values in brackets indicate the numbers of categories. \textbf{c}, Receiver operating characteristic (ROC) curves for the breast cancer detection task (breast cancer vs. non-cancerous/other conditions). \textbf{d-f}, Detailed analysis of the 10-class histological diagnosis task: confusion matrix \textbf{(d)}, weighted F1-scores across models \textbf{(e)}, and top-n accuracy \textbf{(f)}. \textbf{g}, Model performance (AUROC) on three diagnostic tasks across the combined external validation cohorts. \textbf{h}, Representative heatmap visualizations of BRIGHT on three different histological subtypes. The error bars denote the two-sided 95\% confidence interval computed via 1,000 bootstrap resampling. {We demonstrate the $\Delta$ AUC (BRIGHT vs. best or second-best) and p‑value (paired Wilcoxon test) on top of each task in the bar chart.} Can. Det., Cancer detection. Histo. Diag., Histological diagnosis.}
	\label{fig2}
\end{figure}

We first evaluate the diagnostic performance of BRIGHT on, to our knowledge, the most comprehensive and largest curated dataset for breast pathology, comprising a total of 18,250 WSIs for downstream task development, 4,107 WSIs for internal validation and 2,613 WSIs from 5 independent hospitals for external validation (Figure~\ref{fig2}a). The diagnostic benchmark encompasses six clinically essential tasks, including breast cancer detection, multi-class histological subtyping, invasive breast cancer subtyping~(IBC Sub.), Nottingham histologic grading, tumor-infiltrating lymphocytes (TILs) assessment, lymph node metastasis prediction (N-stage), and pathological TNM (pTNM) staging. To further assess generalizability, we also evaluate performance on the publicly available BRACS dataset~\cite{brancati2022bracs}.

BRIGHT demonstrates state-of-the-art (SOTA) diagnostic performance, generally outperforming all benchmarked PFMs across every diagnostic task on the internal validation set (Figure~\ref{fig2}b). For breast cancer detection, BRIGHT achieves an exceptional AUC of 0.992 (95\%CI: 0.989-0.994) on the natural-distribution cohort (Figure~\ref{fig2}c), confirming its robustness and potential as a clinically viable tool to support primary screening and reduce pathologist workload with high reliability. For the more challenging multi-class histological diagnosis task, BRIGHT outperforms SOTA PFMs. It achieves an AUC of 0.973 (95\%CI: 0.966-0.979), a weighted F1-score of 0.901 and a top-1 accuracy of 0.899 (95\%CI: 0.889-0.909) (Figure~\ref{fig2}e-f). Notably, BRIGHT shows outstanding classification performance for high-prevalence categories (Figure~\ref{fig2}d) such as invasive breast carcinoma (IBC), ductal carcinoma in situ (DCIS) and papillary neoplasms (PN), while providing accurate and interpretable attention heatmaps that align with pathological diagnostic criteria (Figure~\ref{fig2}h). The combination of high accuracy and explainability is critical for reliable triage and diagnostic support in routine clinical practice. {Furthermore, we performed further subtyping of IBC based on the real-world natural distribution, categorizing IBC by incidence into invasive ductal carcinoma~(IDC), invasive lobular carcinoma~(ILC), and other subtypes with low occurrence rate. Our method achieved a superior AUC (0.897 (95\%CI: 0.874-0.918)) for IBC subtyping, which facilitates more accurate prognosis assessment, treatment guidance, and recurrence risk prediction.}  Across the remaining diagnostic spectrum, BRIGHT achieves AUCs of 0.898 (95\%CI: 0.884-0.911) for Nottingham histologic grading, 0.967 (95\%CI: 0.950-0.982) for TILs assessment, 0.683 (95\%CI: 0.661-0.704) for lymph node metastasis prediction and 0.661 (95\%CI: 0.640-0.682) for pTNM staging. BRIGHT also outperforms other PFMs on the 7-class histological diagnosis on BRACS with only 547 WSIs (394 WSIs for model training), which demonstrates the capability on limited training samples (Figure~\ref{fig2}b). {The model's generalization is further validated on three independent external cohorts, confirming its generalizability across diverse clinical and institutional settings (Figure~\ref{fig2}g).} Comprehensive performance metrics for each model and diagnostic tasks are provided in Supplementary Table~S6-S8. 

\subsection*{BRIGHT advances molecular subtyping and biomarkers prediction}
Beyond histomorphological diagnosis, the prediction of molecular biomarkers and subtypes from routine H\&E slides represents a transformative application of computational pathology. We rigorously evaluated BRIGHT alongside benchmark PFMs on a comprehensive panel of seven immunohistochemistry (IHC) biomarkers, including estrogen receptor (ER), progesterone receptor (PR), human epidermal growth factor receptor 2 (HER2), Ki-67, androgen receptor (AR), cytokeratin 5/6 (CK5/6), and epidermal growth factor receptor (EGFR). Reflecting recent therapeutic advances, our evaluation includes a clinically critical three-tier HER2 classification task (HER2-zero, HER2-low, and HER2-positive), motivated by evidence that trastuzumab deruxtecan (T-DXd) improves overall and progression-free survival in HER2-low patients compared to standard chemotherapy~\cite{modi2022trastuzumab,wong2025ai}. Furthermore, given the growing importance of immunotherapy, we also evaluate the prediction of programmed death-ligand 1 (PD-L1) expression, a key biomarker for immune checkpoint inhibitor response~\cite{cortes2022pembrolizumab}. Finally, we assess the model's ability on the prediction of the four major molecular subtypes (Luminal A, Luminal B, HER2-enriched, and Triple-negative breast cancer).

\begin{figure}[htp]
	\centering
	\includegraphics[width=.86\linewidth]{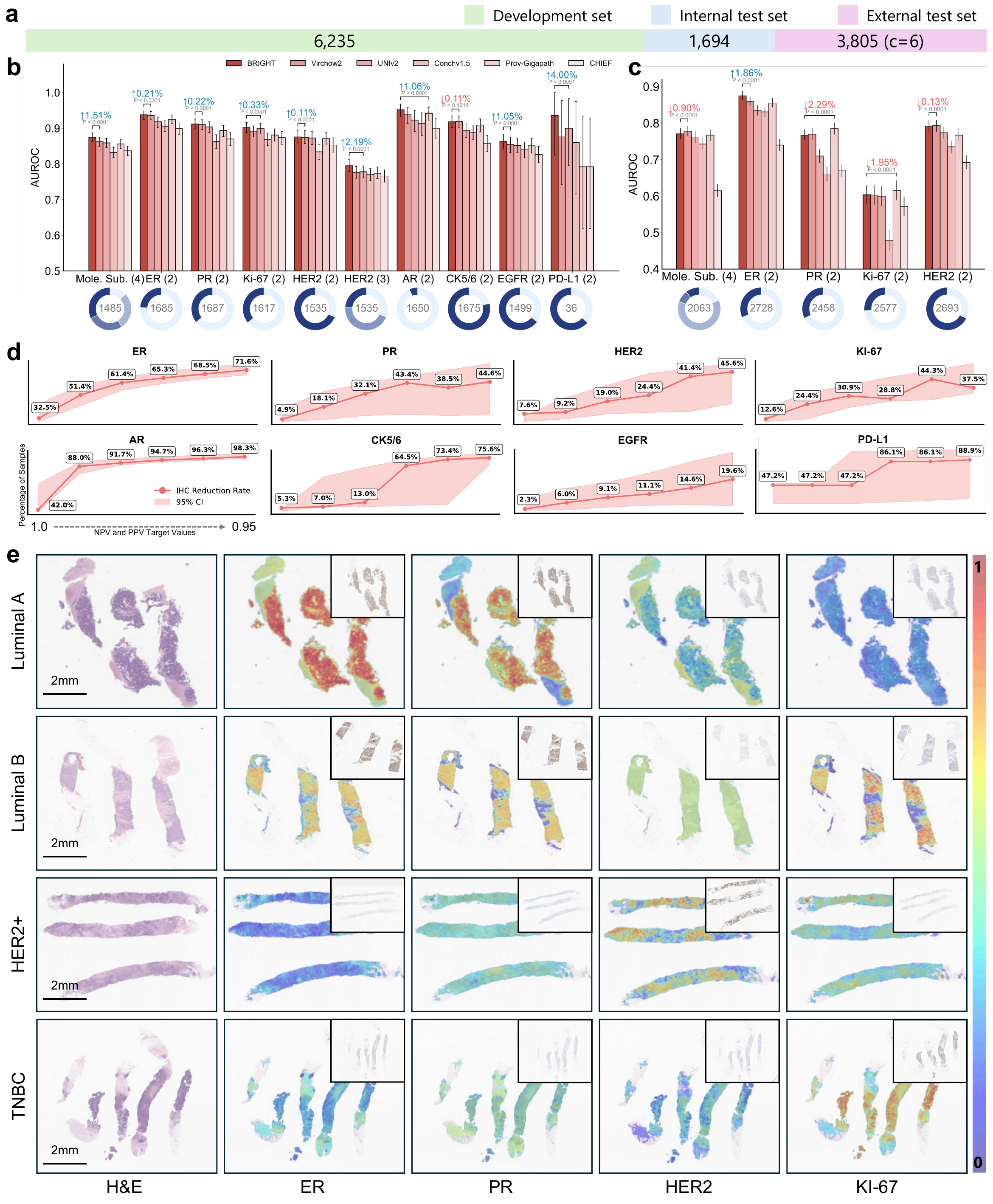}
	\caption{\textbf{Performance of PFMs on molecular subtyping and biomarker prediction tasks of breast pathology.} \textbf{a}, Distribution of the downstream model development, internal test and external test datasets. \textbf{b}, Model performance (AUROC) across individual tasks on the internal validation set. The values in brackets indicate the numbers of categories. \textbf{c}, Model performance (AUROC) on four biomarker prediction tasks and the molecular subtype prediction task across the combined external validation cohorts. \textbf{d}, Potential reduction in immunohistochemistry (IHC) testing, estimated by applying dual clinical thresholds, negative predictive value (NPV) and positive predictive value (PPV), to model predictions. For a given biomarker, samples with a BRIGHT-predicted probability below a predefined NPV threshold are considered true negatives, while those above a PPV threshold are considered true positives. IHC assays for these samples could be potentially waived. The proportion of such samples across the cohort yields the potential IHC assay reduction rate at those specific NPV/PPV thresholds (plotted from 1.0 to 0.95). \textbf{e}, Heatmaps generated by BRIGHT visualize the model's spatial attention when predicting key biomarkers (ER, PR, HER2, Ki-67) across the four major molecular subtypes (Luminal A, Luminal B, HER2-enriched, Triple-negative) of breast cancer. For each comparison, the corresponding diagnostic immunohistochemistry (IHC) image is shown in the top-right inset for direct visual comparison. Shaded areas and error bars denote the 95\% confidence interval computed via 1,000 bootstrap resampling. {We demonstrate the $\Delta$ AUC (BRIGHT vs. best or second-best) and p‑value (paired Wilcoxon test) on top of each task in the bar chart.}}
	\label{fig3}
\end{figure}

On the large-scale internal validation set (Figure~\ref{fig3}a), BRIGHT achieved the highest AUROC in most of the biomarker prediction tasks, attaining an average AUC of 0.938 (95\%CI: 0.925-0.948) for ER, 0.912 (95\%CI: 0.898-0.926) for PR, 0.902 (95\%CI: 0.888-0.916) for Ki-67, 0.876 (95\%CI: 0.859-0.894) for HER2 (binary classification), 0.795 (95\%CI: 0.779-0.812) for HER2 (three-tier classification), 0.952 (95\%CI: 0.933-0.968) for AR, 0.918 (95\%CI: 0.902-0.934) for CK5/6, and 0.863 (95\%CI: 0.842-0.884) for EGFR (Figure~\ref{fig3}b). Furthermore, BRIGHT demonstrated superior performance in predicting the integrated molecular subtype compared to all benchmarked PFMs with AUC of {0.875 (95\%CI: 0.863-0.887)}. Notably, even for the challenging task of PD-L1 prediction with only 289 training samples, BRIGHT achieves a higher AUC of 0.936 (95\%CI: 0.825-1.000). This result underscores the model’s data-efficient nature, which is a critical attribute for translating biomarker prediction tools to clinical settings where large-scale cohorts may not always be available. Comprehensive performance metrics for each model and related tasks on internal test set are provided in Supplementary Table~S9. 

Additionally, based on the integrated molecular subtype prediction task and its required four key biomarkers (ER, PR, HER2, Ki-67), we conduct an independent external validation experiment to rigorously assess the model's generalizability. This combined external cohort comprises data from 6 independent medical centers and includes the publicly available TCGA-BRCA dataset, ensuring diverse clinical settings. BRIGHT demonstrates outstanding model generalizability with highly stable classification performance comparing with the baseline models (Figure~\ref{fig3}c), demonstrating its robustness across diverse and real-world clinical settings. Interpretability analysis reveals that BRIGHT's attention maps show high visual concordance (Figure~\ref{fig3}e) with the expression of corresponding IHC across four key biomarkers for each molecular subtype, greatly improving the trustworthiness of BRIGHT's prediction. Comprehensive performance metrics for each model and related tasks on external test set are provided in Supplementary Table~S10. 


{To directly quantify the potential clinical utility of BRIGHT for biomarker prediction, we estimated its capacity to reduce the need for IHC assays in two specific scenarios. First, in resource-limited settings where IHC is not always immediately available, an H\&E-based prediction model could serve as a pre-screening or triage tool to identify patients with high-probability biomarker positivity, prioritizing them for confirmatory IHC or referral. Second, in high-volume academic laboratories, the same approach could act as a pre-screening filter, reducing the number of routine IHC orders, shortening turnaround time from days to minutes, and freeing resources for more complex cases.} By applying dual clinical thresholds (high Negative Predictive Value, NPV, and high Positive Predictive Value, PPV) to BRIGHT's prediction scores, we identified a substantial subset of cases where IHC testing could be potentially waived with high confidence (Figure~\ref{fig3}d). For instance, at a conservative NPV/PPV threshold of 0.98, BRIGHT could reduce the need for IHC testing by approximately 61.4\% for ER, 32.1\% for PR, 30.9\% for Ki-67, 19.0\% for HER2, 91.7\% for AR, 13.0\% for CK5/6, 9.1\% for EGFR and 47.2\% for PD-L1. Even under the most stringent threshold of 1.0, BRIGHT still achieves a notable reduction of approximately 2.3\%–47.2\% across these biomarkers. {These reduction rates suggest that a substantial proportion of IHC tests could be avoided without compromising diagnostic accuracy, thereby accelerating the conventional diagnostic pathway and reducing associated costs. BRIGHT offers a complementary, cost‑saving pre‑screening solution in settings where IHC is either unavailable or a limited resource. Prospective implementation studies are needed to validate these theoretical reductions in real‑world workflows.}

\subsection*{BRIGHT can predict therapeutic response}
Neoadjuvant therapy (NAT) represents a cornerstone of breast cancer management, aiming to downstage tumors prior to surgery and potentially obviate the need for extensive resection for the cases of pathological complete response (pCR). However, predicting patient response to NAT from pre-treatment histology represents a major yet unmet goal in precision oncology for breast cancer. We evaluated the efficacy of BRIGHT on predicting NAT outcomes using diagnostic H\&E biopsy slides obtained prior to treatment.

\begin{figure}[tp]
	\centering
	\includegraphics[width=.95\linewidth]{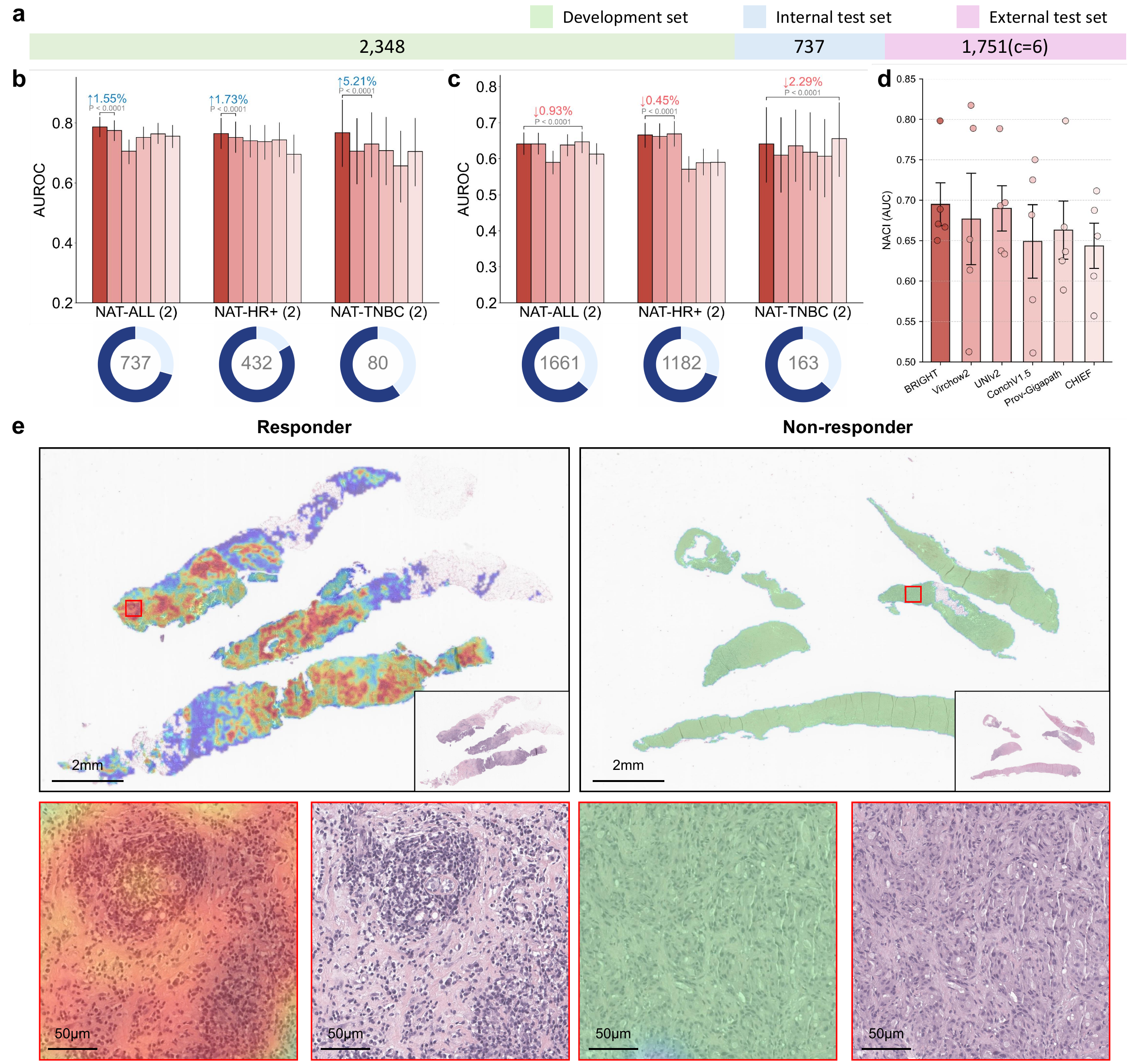}
	\caption{\textbf{Performance of PFMs in predicting breast cancer treatment response.} \textbf{a}, Distribution of the downstream model development, internal test and external test datasets. \textbf{b-c}, Model performance (AUROC) across individual neoadjuvant therapy (NAT) response prediction tasks on the internal \textbf{(b)} and external validation sets \textbf{(c)}. The values in brackets indicate the numbers of categories. \textbf{d}, Performance (AUROC) of the models in predicting response to neoadjuvant immunotherapy for triple-negative breast cancer (TNBC) patients, evaluated using five-fold cross-validation. \textbf{e}, Heatmaps generated by BRIGHT for a patient who responded to immune checkpoint blockade (ICB; left) and a non-responder (right). The model’s attention is prominently localized to regions rich in tumor-infiltrating lymphocytes (TILs), aligning with known immunological correlates of treatment response. Error bars denote the 95\% confidence interval computed via 1,000 bootstrap resampling. {We demonstrate the $\Delta$ AUC (BRIGHT vs. best or second-best) and p‑value (paired Wilcoxon test) on top of each task in the bar chart.}}
	\label{fig4}
\end{figure}

On the held-out internal validation cohort, BRIGHT achieves superior performance in predicting pCR to neoadjuvant therapy (AUC: 0.787 (95\%CI: 0.753-0.820)) compared to all benchmarking PFMs (Figure~\ref{fig4}b). This predictive advantage is generally maintained on the independent multi-center external validation set (AUC: 0.641 (95\%CI: 0.611-0.673)), Figure~\ref{fig4}c), demonstrating robust generalizability. Since neoadjuvant therapeutic strategies for breast cancer are highly subtype-specific. For example, HER2-positive breast cancer requires a dual-blockade strategy combining trastuzumab and pertuzumab with chemotherapy~\cite{loibl2024early}, whereas the standard of care for high-risk triple-negative breast cancer (TNBC) has shifted towards platinum-containing chemotherapy combined with immune checkpoint inhibitors~\cite{schmid2022event}. Thus, we leveraged our large-scale cohort (Figure~\ref{fig4}a) to perform subgroup-specific analyses, which allows us to evaluate model performance within clinically distinct populations that receive standardized, subtype-specific treatment. BRIGHT achieves an AUC of 0.768 (95\%CI: 0.654-0.878) for TNBC, 0.765 (95\%CI: 0.714-0.817) for HR-positive on the internal validation set, 0.641 (95\%CI: 0.534-0.743) for TNBC, 0.666 (95\%CI: 0.631-0.699) for HR-positive on the external validation cohort. In addition, we specifically evaluate BRIGHT on the challenging task of predicting response to neoadjuvant immunotherapy. Despite being trained on an extremely limited cohort of only 180 samples, BRIGHT demonstrated relatively more stable predictive performance compared to benchmark PFMs (Figure~\ref{fig4}d), achieving an average AUC of 0.695. Interpretability analysis further revealed that BRIGHT’s predictions are grounded in the histomorphology of tumor immune microenvironment (TIME). For the immune checkpoint blockade (ICB) responder, BRIGHT identifies immune-hot phenotype with highlighted rich TILs. While for the ICB non-responder, BRIGHT captures immune‑cold phenotype with sparse TILs (Figure~\ref{fig4}e). This finding aligns closely with established clinical evidence indicating that patients exhibiting a more active TIL‑rich tumor‑immune microenvironment are more likely to benefit from immunotherapy~\cite{el2021tale}. The model’s ability to visually localize the TIME features directly from H\&E slides provides a cost‑effective and trustworthiness solution for patient stratification, potentially guiding treatment selection in clinical practice. Comprehensive performance metrics for each model and therapeutic response prediction tasks are provided in Supplementary Table~S11-S12. 

\subsection*{{BRIGHT provides complementary prognostic information beyond established clinicopathological factors}}
Predicting long-term clinical outcomes, such as overall survival (OS) and disease-free survival (DFS), from histopathology images remains a challenging yet clinically important task in breast oncology. {We therefore investigated whether BRIGHT could extract prognostically relevant information from diagnostic H\&E slides and support survival risk stratification across diverse breast cancer settings.}

\begin{figure}[htp]
	\centering
	\includegraphics[width=.95\linewidth]{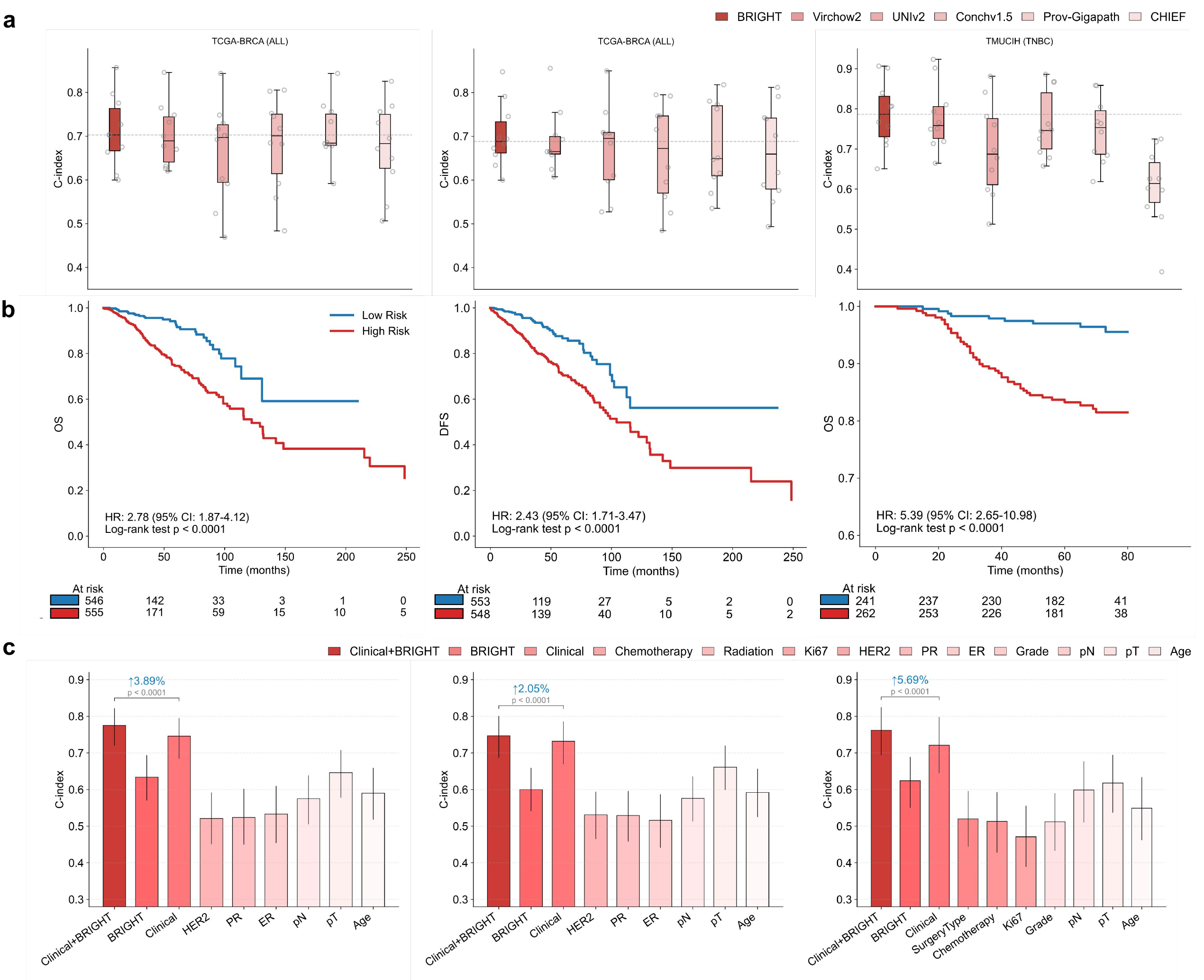}
	\caption{
		\textbf{Performance of PFMs in prognostic prediction.} \textbf{a}, Box plots comparing the C-index of BRIGHT with PFMs across three experimental settings: C26.TCGA-BRCA overall survival (OS), C26.TCGA-BRCA disease-free survival (DFS), and OS in the TNBC cohort of the C1.TMUCIH. Each box summarizes the C-index values obtained from all validation folds in 10-fold cross-validation. \textbf{b}, Kaplan-Meier survival curves for OS and DFS based on BRIGHT-derived risk stratification in the validation sets. Patients were stratified into high-risk (H, red) and low-risk (L, blue) groups according to the median risk score calculated from the training set. Statistical significance was assessed using a two-sided log-rank test. \textbf{c}, {Quantification of the incremental prognostic value of BRIGHT beyond established clinicopathological factors. C-index values are shown for individual clinicopathological variables, clinical-only models, and clinical + BRIGHT models across three survival evaluation settings. The incorporation of BRIGHT consistently improved prognostic discrimination beyond the corresponding clinical models. Blue values indicate the relative improvement ($\%$) in C-index after adding BRIGHT to the corresponding clinical model. P values were calculated using two-sided Wilcoxon signed-rank tests based on 1,000 bootstrap resamples.}}
	\label{fig5}
\end{figure}

The prognostic performance of BRIGHT was first evaluated on the entire C26.TCGA-BRCA cohort for both OS and DFS. Under a 10-fold cross-validation setting, BRIGHT consistently outperforms all PFMs, achieving a concordance index (C-index) of 0.711 for OS and 0.702 for DFS (Figure~\ref{fig5}a, Supplementary Table S14). Patients are subsequently stratified into high-risk and low-risk groups using the median risk score derived from the training set. Patients classified as high risk exhibit significantly worse survival outcomes compared with those in the low-risk group (Figure~\ref{fig5}b). Together, these results demonstrate that BRIGHT effectively captures survival-associated information from H\&E slides and enables robust prognostic stratification in breast cancer. The prognostic performance of BRIGHT was then assessed in the TNBC cohort from C1.TMUCIH. Consistent with the findings in the overall cohort, BRIGHT achieves strong risk stratification in the TNBC cohort, yielding a C-index of 0.787 for OS in the validation set.

{To investigate whether BRIGHT provides prognostic information independent of established clinicopathological factors, we performed multivariable Cox regression analyses using available clinicopathological covariates in each cohort. BRIGHT remained significantly associated with survival outcomes after adjustment for these covariates across all survival evaluation settings, supporting the prognostic relevance of BRIGHT beyond available clinicopathological covariates (Supplementary Table S15-17).}

{Finally, the added prognostic value of BRIGHT beyond conventional clinicopathological assessment was evaluated. Figure~\ref{fig5}c compares the prognostic performance of individual clinicopathological variables, BRIGHT, clinical-only models, and clinical + BRIGHT models across all survival evaluation settings. While models integrating multiple clinicopathological factors substantially outperformed individual variables, incorporation of BRIGHT consistently improved prognostic discrimination beyond the corresponding clinical models. In TCGA-BRCA, the C-index increased from 0.746 to 0.775 for OS and from 0.732 to 0.747 for DFS. Similarly, in the TMUCIH TNBC cohort, incorporation of BRIGHT improved the C-index from 0.721 to 0.762 (Supplementary Table S18). These gains were consistently observed across all survival evaluation settings despite the strong baseline performance of the corresponding clinical models. Together, these findings suggest that BRIGHT captures prognostic information complementary to established clinicopathological factors and may provide additional prognostic value when integrated with conventional clinical assessment.}

\subsection*{{Ablation study on generalist-specialist collaborative design}}
{To evaluate the effectiveness of generalist-specialist collaborative design, we compared three configurations using the same downstream task setup: (1) Generalist‑only (the frozen Virchow2 encoder), (2) Specialist‑only (BRIGHT (S), obtained by LoRA fine‑tuning of Virchow2 on breast pathology data), and (3) Collaborative (BRIGHT, formed by concatenating generalist and specialist embeddings).}

{Across all task domains (diagnosis, biomarker prediction, treatment response, and survival), a consistent pattern emerged. Specialist‑only achieved performance comparable to, or slightly better than, the Generalist‑only baseline~(Supplementary Table S13 and Figure S2). This indicates that LoRA‑based domain adaptation successfully transfers generalist histomorphological knowledge to the breast pathology domain without significant degradation. Collaborative consistently outperformed both Generalist‑only and Specialist‑only configurations in the vast majority of tasks. These results demonstrate that the collaborative integration of complementary feature spaces, combining the broad histomorphological primitives of a generalist PFM with the breast‑specific discriminative patterns learned by a specialist encoder, is the key driver of BRIGHT's superior performance. Neither component alone achieves the same level of accuracy and robustness. All detailed results can be found in the Supplementary Table S13.}

\section*{Discussion}
In this study, we introduce BRIGHT, the first collaborative generalist-specialist pathology foundation model (PFM) specifically designed for breast oncology. Our work addresses a critical gap in computational pathology that lack of organ-specific PFM capable of handling the full clinical spectrum within a specialized domain. By curating the largest multi-institutional benchmarking cohorts for breast pathology to date and employing a novel dual-pathway architecture, we demonstrated that BRIGHT not only sets a new state-of-the-art, outperforming leading generalist PFMs (UNIv2~\cite{chen2024towards}, Virchow2~\cite{vorontsov2024foundation}, CONCHv1.5~\cite{lu2024visual}, {Prov-GigaPath~\cite{xu2024whole} and CHIEF~\cite{wang2024pathology}}), across a comprehensive spectrum of 25 internal and 11 external clinical tasks but also establishes the collaborative generalist-specialist paradigm as a feasible solution for developing high-fidelity organ-specific AI. This paradigm moves the CPath field a step forward that transforming the generalist PFM into a specialist domain and achieving comprehensive, high-level performance within a single, complex organ system. 

BRIGHT’s superior performance across diagnostic, predictive, and prognostic tasks underscores the intrinsic value of integrating domain-specific knowledge. {Given the large scale of our downstream training sets and the real‑world, unselected internal validation cohort, all benchmarked PFMs already achieve high and stable performance. Under this challenging setting, the absolute performance gains of BRIGHT are modest but statistically significant and consistent across tasks, reflecting consistent and robust improvement rather than marginal or unstable gains.} In core diagnostic tasks from cancer detection to complex multi-class subtyping to invasive breast carcinoma subtyping, BRIGHT achieved exceptional accuracy on a natural‑distribution validation cohort, highlighting its potential for reliable clinical decision support and workload reduction. {More significantly, its ability to predict key molecular biomarkers (including the challenging three‑tier HER2 classification and PD‑L1 status) directly from H\&E slides could serve a practical adjunctive role in specific clinical scenarios. For instance, in resource‑limited settings where IHC is not always immediately available, or in high‑volume laboratories seeking to optimize workflow, BRIGHT could function as a pre‑screening or triage tool to identify high‑confidence predictions, thereby potentially reducing the number of routine IHC orders. Prospective implementation studies are needed to validate whether these theoretical reductions translate into real‑world workflow improvements and cost savings.}

{Here, we also performed exploratory analyses to evaluate the potential of BRIGHT for treatment response and survival prediction. Our results show that BRIGHT provides subtype‑stratified predictions for neoadjuvant therapy response with moderate performance, although substantial improvement is still needed for clinical application. Its capacity to identify “immune‑hot” versus “immune‑cold” tumor immune microenvironments from routine H\&E slides aligns with established biological principles~\cite{el2021tale} and suggests a potential direction for future exploration in immunotherapy patient selection. 
	{Furthermore, BRIGHT-derived risk scores provided prognostic information complementary to established clinicopathological variables in our retrospective survival analyses. The observed improvements in prognostic performance when BRIGHT was integrated with clinical models suggest that BRIGHT may capture information not fully represented by routinely assessed clinicopathological factors.}}

The collaborative architecture of BRIGHT directly tackles the limitations of current generalist PFMs by harmonizing breadth with depth. While multi-organ pretraining provides a robust foundation of generalist histomorphological patterns, it may dilute subtle, organ-specific diagnostic features critical for specialist interpretation. BRIGHT resolves this by maintaining a frozen, well-pretrained generalist encoder (Virchow2~\cite{vorontsov2024foundation,zimmermann2024virchow2}) to preserve broad knowledge, while a parallel and efficiently fine-tuned specialist pathway (via LoRA~\cite{hu2022lora}) deepens expertise in breast-specific pathology. This design enables complementary feature integration, where the generalist pathway captures universal representations while the specialist pathway refines domain-specific discriminative patterns, thereby enhancing overall representational capacity. Moreover, this approach is highly parameter- and data-efficient, as it leverages well-trained PFM from millions of WSIs and thus avoids the prohibitive cost of training a comparably outstanding specialist model entirely from scratch. This algorithmic design is validated not only by BRIGHT’s outperformance of all benchmark generalist PFMs but also by the superior interpretability of its attention maps, which align accurately with known pathological landmarks and biomarker expression patterns. This collaboration between generalist robustness and specialist precision is a core conceptual contribution of our work.

Despite its strengths, our study has several limitations. First, while our cohorts are the largest among current PFM studies, they are still retrospective. Prospective, multi-center clinical trials are needed to definitively establish BRIGHT’s impact on diagnostic accuracy, workflow efficiency and patient outcomes. Second, the current model is designed for H\&E-stained slides. Integrating multimodal data, such as immunohistochemistry, genomic profiles and pathology report~\cite{xu2025multimodal,yan2025multimodal} within the foundation model framework could enhance even more comprehensive representation and knowledge and expand the model capability to cross-modal tasks like visual question answering~\cite{he2020pathvqa} and high-dimensional spatial-omics generation~\cite{li2026ai,valanarasu2025multimodal}. {Third, although BRIGHT demonstrated higher robustness than generalist PFMs in external validation, the performance gains were less pronounced than those observed in internal validation. This discrepancy likely reflects domain shifts across institutions, including variations in tissue fixation, staining protocols, scanner characteristics, and patient populations. Future works should focus on developing effective domain adaptation, test-time generalization, or federated learning approaches to further close the internal–external performance gap while preserving patient privacy. Finally, we acknowledge that for therapy response and survival prediction tasks, the current performance levels remain exploratory and are not yet sufficient for standalone clinical decision-making. These tasks will require further methodological advances and dedicated prospective validation.}

In conclusion, BRIGHT demonstrates the potential value of a collaborative generalist-specialist paradigm to meet the diverse and specialized clinical demands of breast pathology. Through large-scale pre-training on breast histology and comprehensive validation across diagnostic, predictive, and prognostic tasks, the model delivers robust, expert-level performance. Our study, combining systematic multi-institutional data curation, a novel dual-pathway architecture and rigorous clinical benchmarking, provides a delightful direction for developing high-fidelity, organ-specific foundation models in oncology. These findings highlight a promising pathway for advancing next-generation PFM in other medical specialties by transitioning from generalist breadth to specialist depth. {While further prospective validation is needed before clinical deployment, this work lays a foundational step toward delivering precise and personalized care, ultimately lighting a ‘BRIGHT’ path for patients with breast cancer.}

\section*{Methods}

\subsection*{{Million-scale training dataset for BRIGHT pre-training}}
All data were de‑identified prior to computational analysis and model development. This retrospective study did not involve direct patient recruitment or prospective intervention and was approved by the Institutional Review Boards of all participating centers. Informed consent was waived in accordance with institutional policies for research using de‑identified archival specimens. To train the breast-specific pathology foundation model (BRIGHT), we construct a large-scale cohort of breast pathology whole slide images (WSIs) in this study. The foundation model training dataset comprises over 210 million non-overlapping breast pathology tiles extracted from 51,836 H\&E stained WSIs from 40,203 patients, sourced from 19 independent clinical centers. The details of data collection of multiple centers are given in Supplementary Table S5. It is predominantly composed of invasive breast carcinomas, supplemented by a wide spectrum of breast lesions (e.g., fibroadenoma, usual ductal hyperplasia). All slides were digitized at $\times$20 (0.5 mpp) or $\times$40 (0.25 mpp) magnification.

\subsection*{BRIGHT architecture and training}
BRIGHT is built upon a dual-pathway architecture consisting of a generalist encoder and a specialist encoder with identical network structure, Vision Transformer 'huge' architecture (ViT-H/14)~\cite{dosovitskiy2020image}. For the generalist encoder, we apply an open-source PFM (Virchow2~\cite{vorontsov2024foundation,zimmermann2024virchow2}) trained on the largest dataset to date with over 3 million pan-cancer WSIs. To adapt this generalist representation to the domain of breast pathology, we employ the widely used DINOv2~\cite{oquab2023dinov2} as the self-supervised learning strategy and adapt low-rank adaptation (LoRA)~\cite{hu2022lora} as a parameter-efficient fine-tuning strategy, yielding the specialist encoder BRIGHT (S). Specifically, all parameters of the Virchow2 backbone remain frozen, and lightweight LoRA modules are injected into the query, key and value projection matrices of each self-attention layer. These LoRA modules employ a low-rank decomposition with rank 8, enabling efficient domain adaptation by updating only a small number of additional parameters. This design substantially reduces training and storage costs while maintaining the representational capacity of original model. The  number of trainable parameters after LoRA injection is 24.45M with total parameters 655.60M. BRIGHT was trained using 16 NVIDIA A800 GPUs. 

\subsection*{{BRIGHT embedding}}
For an input histopathology tile of size 224$\times$224 pixels, the final BRIGHT embedding was constructed by concatenating the 2560‑dimensional feature vector from the specialist encoder BRIGHT (S) with the 2560‑dimensional feature vector from the generalist Virchow2 encoder, resulting in a 5120‑dimensional representation. The BRIGHT embedding explicitly integrates complementary features with both breast-specific patterns and pan-cancer histomorphological primitives, providing a richer and more discriminative representation for downstream tasks.

\subsection*{{Aggregator for downstream tasks}}
For the downstream task training and evaluation, we apply a widely used multiple-instance learning (MIL) aggregator (CLAM)~\cite{lu2021data} to integrate tile-level features into a slide-level prediction. The aggregator receives the set of tile embeddings from a WSI as input and outputs a task-specific prediction. For each downstream task, the BRIGHT encoders were frozen, and only the aggregator network was trained from scratch using labeled data from the respective benchmarking dataset.

\subsection*{Benchmarking datasets}
To comprehensively evaluate the performance of BRIGHT across the full clinical spectrum of breast pathology, we constructed the largest multi-institutional benchmarking dataset to date, including 24 private cohorts and two public datasets (TCGA-BRCA~\cite{weinstein2013cancer} and BRACS~\cite{brancati2022bracs}). This resource integrates a diverse collection of public and private data, encompassing diagnostic, prognostic, and predictive tasks. A detailed breakdown of sample sizes, slide counts, and associated clinical endpoints for each dataset is provided in Supplementary Table~S1-S2.

The primary cohort for downstream task development was derived from Tianjin Medical University Cancer Institute and Hospital (C1.TMUCIH). To reflect real‑world clinical prevalence, data were consecutively collected without selection based on diagnosis, with exclusions applied only for insufficient image quality or incomplete clinical annotation. To ensure a robust and representative internal evaluation, a held‑out validation cohort was constructed, comprising all consecutive patients undergoing breast biopsy during the full calendar year of 2024 at C1.TMUCIH. This temporal hold‑out strategy preserves the natural, unselected case distribution encountered in routine practice.

For each downstream task within the internal cohort, the corresponding dataset was randomly split at the patient level into an 80\% training set for model development and a 20\% validation set for hyperparameter tuning and model selection. For patients with multiple slides, all slides from the same patient were assigned to the same split to prevent data leakage. Critically, all WSIs in both internal validation and external validation datasets used for downstream task evaluation are strictly held out from the BRIGHT foundation model's pre‑training phase and {the downstream task training phase}, ensuring an unbiased assessment of its generalizable feature representations.

\subsection*{Label extraction}
Clinical and pathological labels for diagnostic and biomarker prediction tasks, including breast cancer detection, histopathological diagnosis, {invasive breast cancer subtyping (IBC Subtyping)}, Nottingham grade, tumor‑infiltrating lymphocyte (TIL) density, pTNM stage, biomarker status (ER, PR, HER2, etc.), Molecular subtype, were extracted directly from standardized pathology reports. Treatment response labels for neoadjuvant therapy (NAT) and neoadjuvant immunotherapy (NIT) were defined based on postoperative pathological assessment of pathological complete response (pCR vs. non‑pCR). Long‑term survival endpoints, including overall survival (OS) and disease‑free survival (DFS), were obtained from maintained clinical follow‑up records. Detailed definitions for each label, including specific diagnostic criteria, biomarker thresholds, and outcome measures, are provided in Supplementary Table~S4. {Notably, the categories for both the histopathological diagnosis task and the invasive breast carcinoma (IBC) subtyping task were derived from the real-world distribution of consecutive breast core-needle biopsies performed at C1.TMUCIH between 2019 and 2024~(Supplementary Figure S1). For the histopathological diagnosis task, all diagnostic entities with more than 100 patients were assigned individual classes, while the remaining rare entities were grouped into an ‘Others’ class. ‘Normal’ was defined as cases dominated by normal breast tissue without a specific pathological lesion. ‘Benign non-neoplastic lesions’ included benign reactive, inflammatory, fibrotic, or mixed non-neoplastic changes without a single dominant retained diagnostic entity. ‘Others’ consisted of specific low-prevalence diagnostic entities that did not reach the predefined sample-size threshold for separate modeling. For specific benign lesions such as UDH and fibroadenoma, the named entity was assigned as the class label only when it constituted the dominant histological diagnosis. For the IBC subtyping task, samples were classified into invasive ductal carcinoma (IDC), invasive lobular carcinoma (ILC), and Others.}

\subsection*{Statistical analysis}
Model performance for diagnostic, biomarker prediction, and treatment response tasks was evaluated using the area under the receiver operating characteristic curve (AUROC). Survival prediction performance was assessed with the concordance index (C‑index). All performance metrics were reported based on 10-fold cross-validation.
For survival analyses, patients were stratified into low‑ and high‑risk groups based on the median risk score derived from the training set. Survival probabilities were estimated using the Kaplan‑Meier method, and differences between risk groups were compared with the log‑rank test. Hazard ratios (HRs) and corresponding 95\% CIs were calculated using  multivariable Cox proportional hazards models, as indicated in the respective results sections.

\subsection*{Authors' Disclosures}
The authors declare no competing interests.

\subsection*{Authors' contributions}
X.G. and Z.L. contributed to study conception and coordination, building a cross-center partnership and research team and getting data sharing agreements.
C.H. and J.L. developed infrastructure and trained models throughout the study worked on drafting and revising the manuscript.
J.Z., M.Z., F.M., N.M. and F.L. revised the manuscript.
Y.J. and J.H. performed evaluation and analysis.
Z.X. and W.L. contributed as research advisors.
L.W.,
J.C.,
Q.L.,
W.W.,
L.C.,
W.Y.,
Z.C.,
X.W.,
J.Y.,
X.J.,
B.L.,
H.S.,
J.L.,
C.L.,
Y.C.,
Y.D.,
X.Q.,
L.Y.,
Y.Y.,
G.G.,
Y.L.,
Y.Y.,
L.Z.,
C.L.,
T.Z.,
Z.S.,
C.L.,
X.Z. and J.X. worked on data preparation.

\subsection*{Acknowledgements}
This work was funded 
Noncommunicable Chronic Diseases-National Science and Technology Major Project (2024ZD0531100, 2024ZD0531101); 
Key Program of the National Natural Science Foundation of China (82530070); 
Tianjin Key Medical Discipline Construction Project (TJYXZDXK-3-016C);
Tianjin Medical University Cancer Institute \& Hospital Medical Technology Innovation Capability Support and Enhancement Project (CXZC-YS202502);
National Natural Science Foundation of China (82572321 and 82372044); 
Natural Science Foundation for Distinguished Young Scholars of Guangdong Province (2023B1515020043);
China Postdoctoral Science Foundation (2025M772908);
National Science Foundation for Young Scientists of China (82502464 and 82402228). 

\subsection*{Data availability}

{Publicly available data used in this study include the TCGA-BRCA dataset, accessible via the NIH Genomic Data Commons (\url{https://portal.gdc.cancer.gov}), and the BRACS dataset, accessible via the official repository (\url{https://www.bracs.icar.cnr.it/}). All in-house patient data were retrospectively collected, fully de-identified, and utilized under institutional review board approvals (No. bc20260031). Due to institutional data governance policies, these private datasets are not publicly available. Researchers interested in accessing specific subsets of the in-house data for academic, non-commercial purposes may submit a formal data use proposal to the corresponding author. Such requests will be reviewed on a case-by-case basis to ensure compliance with ethical, privacy and regulatory obligations.}

\subsection*{Code availability}
The source code and the BRIGHT model will be released upon acceptance.

\appendix

\newpage
\bibliography{refs}


\end{document}